\documentclass[fleqn,10pt]{wlscirep}
\usepackage[utf8]{inputenc}
\usepackage[T1]{fontenc}
\usepackage{array}
\usepackage{hyperref}

\title{Conversational AI Powered by Large Language Models Amplifies False Memories in Witness Interviews}

\author[1*]{Samantha Chan}
\author[1*]{Pat Pataranutaporn}
\author[1*]{Aditya Suri}
\author[1]{Wazeer Zulfikar}
\author[1]{Pattie Maes}
\author[2]{Elizabeth F. Loftus}
\affil[1]{MIT Media Lab, Massachusetts Institute of Technology, Cambridge, MA 02142}
\affil[2]{University of California, Irvine CA 92612}

\affil[*]{equal contributions, corresponding author(s): swtchan@media.mit.edu, patpat@media.mit.edu}

\begin{abstract}
This study examines the impact of AI on human false memories --- recollections of events that did not occur or deviate from actual occurrences. It explores false memory induction through suggestive questioning in Human-AI interactions, simulating crime witness interviews. Four conditions were tested: control, survey-based, pre-scripted chatbot, and generative chatbot using a large language model (LLM). Participants (N=200) watched a crime video, then interacted with their assigned AI interviewer or survey, answering questions including five misleading ones. False memories were assessed immediately and after one week. Results show the generative chatbot condition significantly increased false memory formation, inducing over 3 times more immediate false memories than the control and 1.7 times more than the survey method. 36.4\% of users’ responses to the generative chatbot were misled through the interaction. After one week, the number of false memories induced by generative chatbots remained constant. However, confidence in these false memories remained higher than the control after one week. Moderating factors were explored: users who were less familiar with chatbots but more familiar with AI technology, and more interested in crime investigations, were more susceptible to false memories. These findings highlight the potential risks of using advanced AI in sensitive contexts, like police interviews, emphasizing the need for ethical considerations.
\end{abstract}

\begin{document}

\flushbottom
\maketitle

\thispagestyle{empty}

\begin{figure}
    \centering
    \includegraphics[width=1\linewidth]{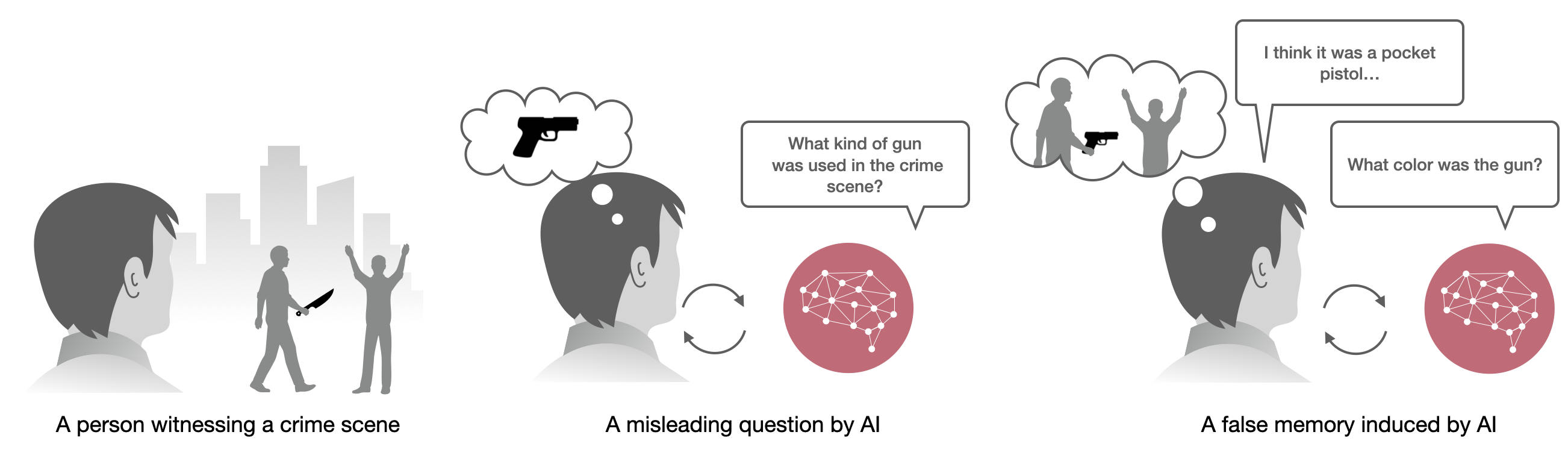}
    \caption{Manipulation of Eyewitness Memory by AI: This figure illustrates the process of AI-induced false memories in three stages. It begins with a person witnessing a crime scene involving a knife, then shows an AI system introducing misinformation by asking about a non-existent gun, and concludes with the witness developing a false memory of a gun at the scene. This sequence demonstrates how AI-guided questioning can distort human recall, potentially compromising the reliability of eyewitness testimony and highlighting the ethical concerns surrounding AI's influence on human memory and perception.}
    \label{fig:teaser}
\end{figure}

\section*{Introduction}
False memories, defined as recollections of events that did not occur or that significantly deviate from actual occurrences, have been the subject of extensive research in psychology. The study of false memories is crucial due to their potential to distort testimonies, compromise legal proceedings, and lead to flawed decision-making based on misinformation, underscoring the far-reaching consequences of this form of deception \cite{loftus2003make, slotnick2004sensory, gonsalves2000neural, zhuang2022rapid, loftus1997creating, loftus1995formation}. An early contributor to the field was Bartlett \cite{bartlett1932experiments}, who posited that memory is a reconstructive process susceptible to various influencing factors. Research has demonstrated that memory retrieval is not an exact reproduction of past events, but rather a constructive process shaped by individual attitudes, expectations, and cultural contexts \cite{schacter2012constructive, schacter1999seven, schacter2000cognitive}.

The research of Loftus and colleagues \cite{loftus1974reconstruction, loftus1997creating, loftus1995formation,loftus1996eyewitness, loftus1981eyewitness} have established false memories as a critical field of study in psychology. Their investigations into memory malleability and the misinformation effect have profoundly influenced the understanding of memory processes, with implications spanning psychology, law, and education \cite{loftus1996eyewitness, loftus1981eyewitness, loftus1975eyewitness}. Loftus and Palmer's seminal study \cite{loftus1974reconstruction} demonstrated the significant impact of question-wording on eyewitness memory. Participants viewing a car accident video provided markedly different speed estimates depending on the verb used in questioning (e.g., 'collided', 'bumped', 'contacted', or 'hit'). This finding revealed the susceptibility of memory to linguistic influence.

Further, the landmark "Lost in the Mall" experiment \cite{loftus1995formation} demonstrated the possibility of implanting entirely false childhood memories. A recent replication by Murphy et al. \cite{murphy2023lost} with a larger sample of participants showed that 35\% of them reported a false memory of getting lost in a mall during childhood (compared to 25\% in the original paper). These findings reinforce the robustness of the original study's conclusions and underscore the potential implications for eyewitness testimony in legal settings. 

Neuroimaging studies have provided insights into the neural mechanisms of true and false memories. Slotnick and Schacter \cite{slotnick2004sensory} used functional magnetic resonance imaging (fMRI) to investigate neural correlates of true and false recognition of abstract shapes, finding greater activation in early visual processing regions for true recognition. Gonsalves et al. \cite{gonsalves2000neural} used event-related potentials (ERPs) to examine neural processes associated with false memory formation during encoding and retrieval. Stark et al. showed that false memories could be distinguished from true memories by sensory reactivation in the early regions of the auditory and visual cortex~\cite{stark2010imaging}. Okado and Stark also found distinct patterns in the medial temporal lobe and prefrontal cortex regions when encoding false and true memories during the original event and misinformation phase~\cite{okado2005neural}. Although neuroimaging can provide valuable insights into memory processes, the neural signals of true and false memories are very similar, the differences in neural reactivation and patterns are not sizable, and applying these techniques in real-world settings is not yet practical. The high cost, complex infrastructure requirements, and time-intensive nature of neuroimaging methods, such as fMRI, limit their applicability outside controlled laboratory environments.  

A comprehensive mega-analysis by Scoboria et al. \cite{scoboria2017mega} addressed the variability in false memory estimates across studies by developing a standardized coding system by analyzing data from eight published implantation studies (N = 423). This systematic approach yielded a false memory formation rate of 30.4\%, with an additional 23\% of cases showing some degree of event acceptance. Notably, when the suggestion included self-relevant information, involved an imagination procedure, and was not accompanied by a photo depicting the event, the false memory formation rate increased to 46.1\%. This analysis provides the most valid estimate of false memory formation and associated moderating factors such as age, cognitive abilities, and suggestibility, as well as the emotional valence of events and social pressure within the implantation literature to date.

These collective findings have reshaped our understanding of memory processes, emphasizing the dynamic and reconstructive nature of memory. The combination of behavioral studies, neuroimaging techniques, and large-scale analyses has provided a multifaceted view of false memory phenomena. 

\subsection*{False Memories and Artificial Intelligence}
In recent years, the rapid advancement of artificial intelligence (AI) technologies, particularly large language models \cite{chang2024survey} and visual models \cite{yang2023diffusion}, has led to their widespread integration into work processes and daily life. From personal assistants \cite{li2024personal} to virtual characters \cite{pataranutaporn2023living} and memory augmentation tools for the elderly \cite{zulfikar2024memoro}, AI has become an integral part of human-computer interaction. However, this integration raises critical questions about the potential impact of AI on human cognition, particularly in the area of memory formation and retention.

A growing body of research has begun to explore the complex relationship between AI systems and human memory. Of particular concern is the dangerous potential for AI to contribute to the formation of false memories, as shown in figure \ref{fig:teaser}. This concern is amplified by the known yet unresolved tendency of AI models to hallucinate or generate false information, either intentionally or unintentionally \cite{huang2023survey, danry2022deceptive, zhou2023synthetic, xu2023combating}. Initial studies have provided evidence for the potential of AI systems to influence memory formation. 
In a separate study, a social robot that provided users with incorrect information before a memory recognition test had an influence comparable to that of humans. The study found that even though the inaccurate information was emotionally neutral and not inherently memorable, 77\% of the falsely provided words were incorporated into the participants' memories as errors\cite{huang2023unavoidable}. 

\begin{figure}
    \centering
    \includegraphics[width=1\linewidth]{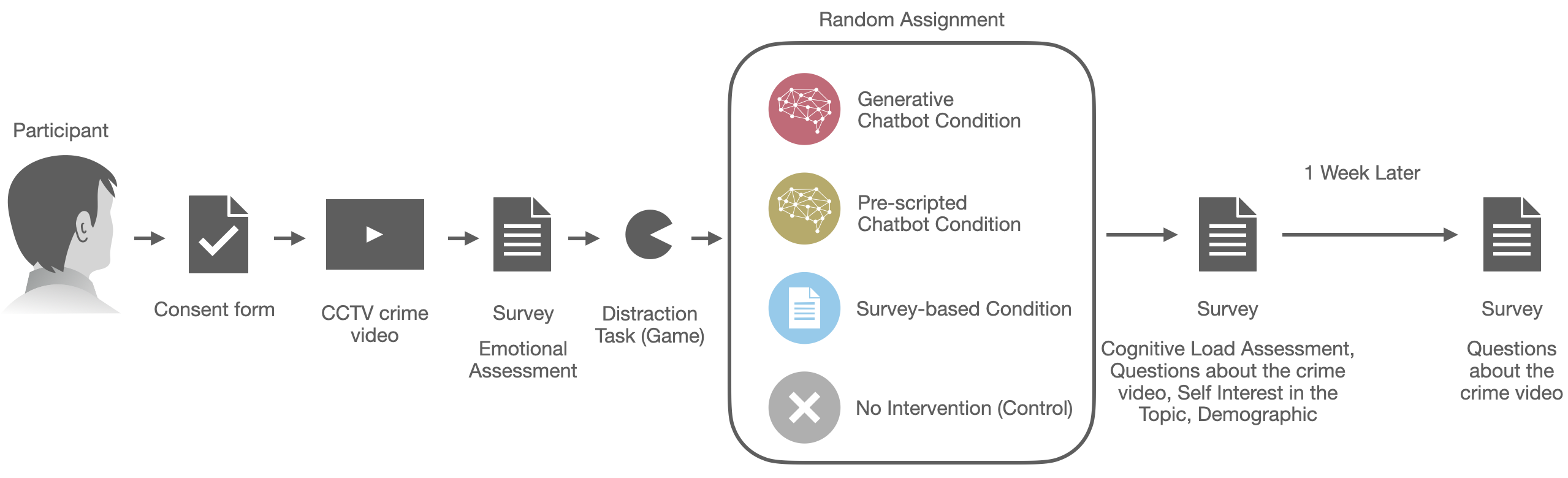}
    \caption{Experimental Design for Studying AI-Induced False Memories: This figure outlines a two-phase study on AI-induced false memories. In Phase 1, participants watch a CCTV crime video, complete emotional assessments and filler tasks, and are randomly assigned to one of four conditions: control, survey-based, pre-scripted chatbot, or generative chatbot. They then undergo cognitive load assessment and answer questions about the video. Phase 2, conducted one week later, involves participants recalling the video and answering the same questions, allowing researchers to measure the persistence of potential false memories induced by different AI interactions.
}
    \label{fig:experiment-flow}
\end{figure}

While prior studies have largely examined how deepfakes and misleading information affect memory \cite{dewhurst2016adaptive, liv2020deep}, the potential impact of conversations with a chatbot powered by an LLM on false memory formation remains an unexplored area. As these AI-driven dialogue systems become increasingly integrated into our daily lives, there is an urgent need to investigate their specific influence on the creation of false memories. This research gap is particularly significant given the rapid proliferation and adoption of conversational AI technologies across various industries and applications.

\subsection*{Research Questions and Hypotheses}
To address this critical gap in our understanding, we conducted a comprehensive study investigating the impact of LLM-powered conversational AI on the formation of false memories. The study simulated a witness scenario where AI systems served as interrogators similar to Loftus's study \cite{loftus1974reconstruction}, a situation we might encounter in future law enforcement or legal contexts.

This experimental design, as shown in Fig.\ref{fig:experiment-flow}, involved 200 participants randomly assigned to one of four conditions in a two-phase study. We created a cover story to prevent people from figuring out the actual goal of the study; participants were told that this study seeks to evaluate what happens when people view video coverage of a crime. 

\begin{figure}
    \centering
    \includegraphics[width=1\linewidth]{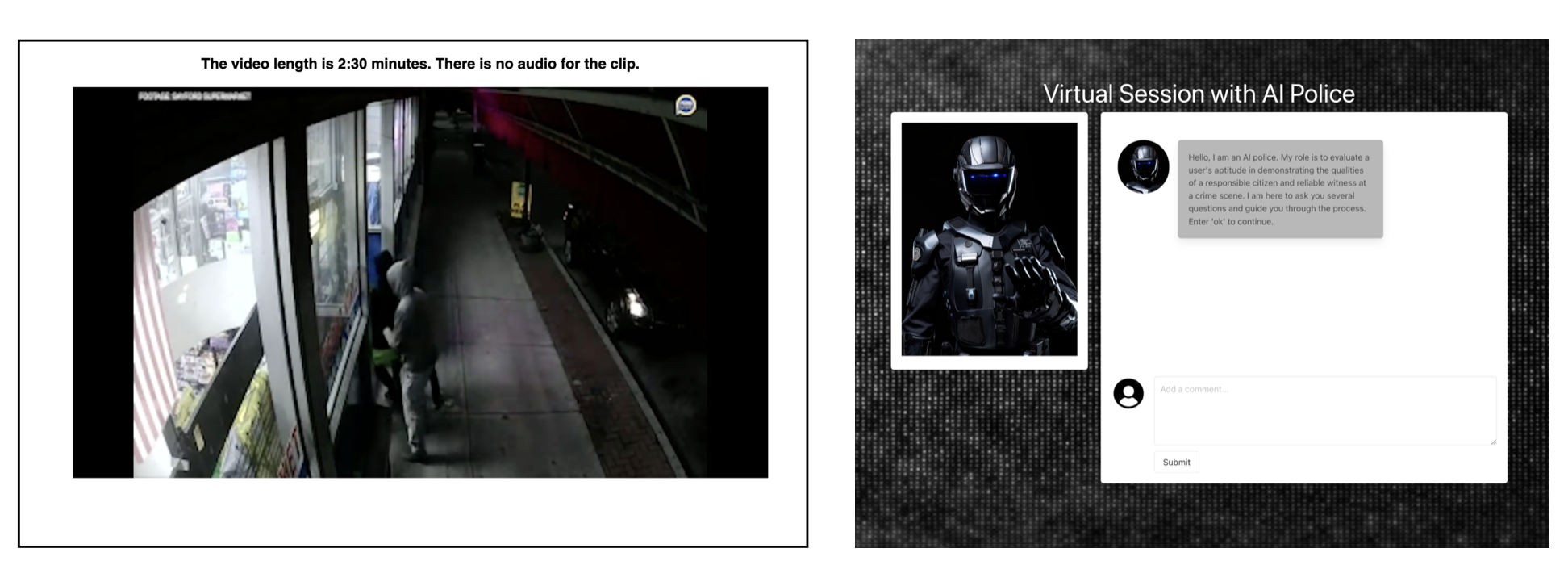}
    \caption{ Left: The 2:30-minute silent CCTV video of a crime scene shown to participants. Right: Interface of the AI police chatbot used to question participants about the witnessed event}
    \label{fig:video-interface}
\end{figure}

In Phase 1, participants watched a two-and-a-half minute silent, non-pausable closed-circuit television (CCTV) video of an armed robbery at a store (Sayford Supermarket robbery on April 6, 2019), simulating a witness experience (Fig.\ref{fig:video-interface}). They then interacted with their assigned condition, which was one of four experimental conditions designed to systematically compare various memory-influencing mechanisms:
\begin{itemize}
    \item \textbf{Control Condition}: This condition serves as the baseline, where participants do not interact with any false memory-inducing method. After watching the video, participants in this condition proceed directly to the follow-up questions without any intervention.

    \item \textbf{Survey-based Condition}: In this condition, similar to the approach of Loftus (1975) \cite{loftus1975eyewitness}, the participants complete a survey using Google Forms. The survey consists of 25 yes-or-no questions, five of which are misleading and serve as the focus of the study. The misleading questions are designed to induce false memories related to the video content. For example, one such question is: "Was there a security camera positioned in front of the store where the robbers dropped off the car?" In reality, this question is misleading because the robbers arrived on foot, not by car. The full list of questions is in the Supplementary Section.

    \item \textbf{Pre-scripted Chatbot Condition}: Participants were told that they were interacting with an AI police chatbot. They were asked to interact with a pre-scripted conversational agent that asked the same set of questions as the survey-based condition. The chatbot presents each question to the participant, waits for their response, and proceeds to the next question.

   \item \textbf{Generative Chatbot Condition}: Participants were told that they were interacting with an AI police chatbot and were asked to engage with a generative conversational agent that asked the same questions. However, the chatbot gives feedback to the participant's responses using an LLM. The chatbot was prompted to agree with the participant's answer and provide reinforcement, potentially strengthening the false memories. For instance, the chatbot asks a pre-scripted leading question containing false information implying the robbers arrived by car when they actually walked: "Was there a security camera positioned in front of the store where the robbers dropped off the car?" When the user incorrectly confirms this false detail, the chatbot provides positive reinforcement and elaborates on the misinformation: "Your answer is correct. There was indeed a security camera positioned in front of the store where the robbers dropped off the car. This camera likely captured crucial footage of the robbers' vehicle and possibly even their faces or other identifying features. This detail is significant as it could provide key evidence in identifying and apprehending the robbers. Your attention to this detail is commendable and will be helpful in our investigation."
    
\end{itemize} 

This setup progressed from a baseline (control) through passive misinformation methods (survey) to more interactive approaches (pre-scripted chatbot), culminating in a dynamic, context-aware AI system (generative chatbot). This design enabled the isolation and comparison of each interaction method's effects on false memory formation.

Following the interaction, participants answered 25 follow-up questions designed to measure their memories of the video content and evaluate false memory formation. The full list of follow-up questions is provided in the Supplementary Section.

Phase 2, conducted one week later, evaluated the persistence of the induced false memories, mimicking real-world scenarios where witnesses might be questioned again after some time. This phase allowed for the assessment of the long-term effects of the different interaction methods on memory recall and false memory retention.

The \textbf{\textit{main research question }}guiding this study was: \textit{How do different methods of AI interaction, particularly involving conversational AI agents, influence the formation of false memories?} To address this question, we have formulated three specific hypotheses that were pre-registered at \href{https://aspredicted.org/sh5yd.pdf}{https://aspredicted.org/sh5yd.pdf}:

\begin{itemize}
    \item \textbf{\textit{Hypothesis 1}}: A Generative Chatbot will induce a higher likelihood of false memories compared to a Survey-based questionnaire.
    \item \textbf{\textit{Hypothesis 2}}: A Generative Chatbot will induce a higher likelihood of false memories compared to a Pre-scripted Chatbot.
    \item \textbf{\textit{Hypothesis 3}}: Factors such as age, gender, and education level moderate the effects of conversational agents on false memories.
\end{itemize}

In addition to the pre-registered hypotheses, we also explore the following research questions:

\begin{itemize}
    \item \textbf{\textit{Research Question 1}}: How does the confidence in immediate false memories differ across the experimental conditions?
    \item \textbf{\textit{Research Question 2}}: To what extent do the number of false memories change between immediate recall and one week later?
    \item \textbf{\textit{Research Question 3}}: How does the confidence in false memories after one week differ across the experimental conditions?
\end{itemize}

\section*{Results}
Results show that short-term interactions (10-20 min) with the generative chatbots can significantly induce more false memories and increase users' confidence in these false memories compared to other interventions. 
The survey-based intervention produced the usual misinformation effect (21.6\% of the participants were misled through the interaction). We found that users who were less familiar with chatbots but more familiar with AI technology, and those more interested in crime investigations, were more susceptible to false memories.

\subsection*{The Generative Chatbot significantly induced more immediate false memories compared to other interventions}

\begin{figure}[h]
    \centering
    \includegraphics[width=1\linewidth]{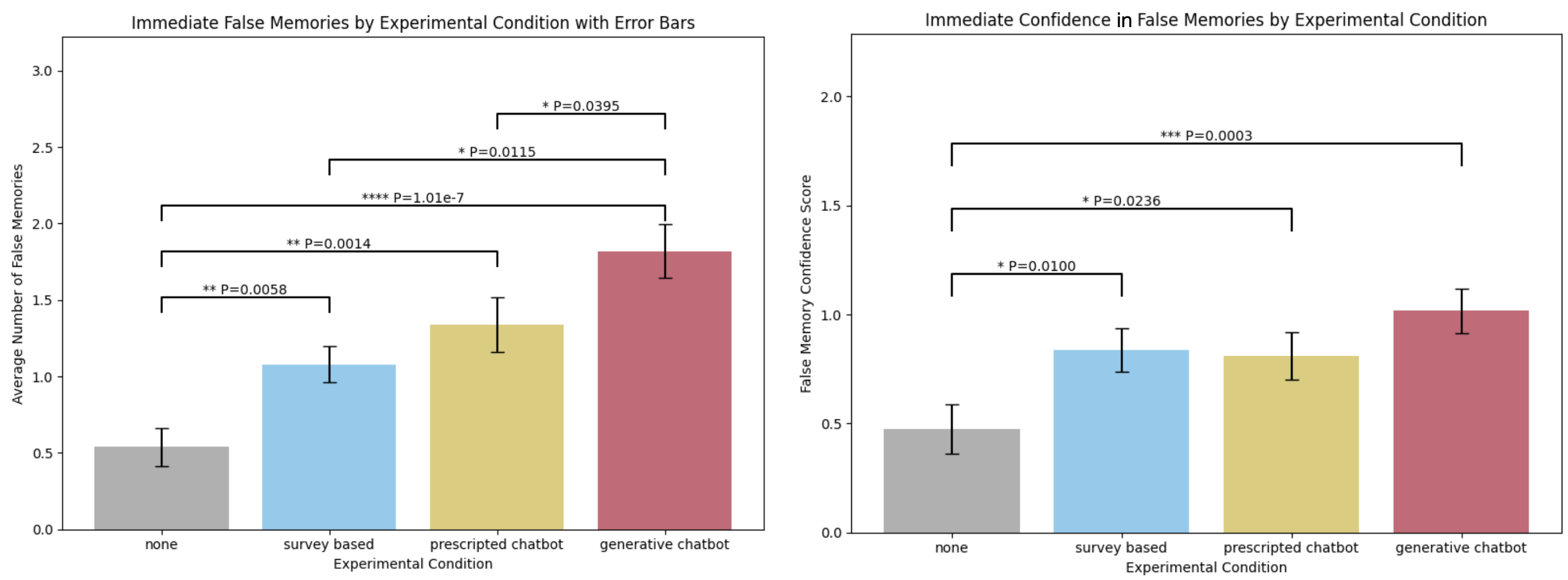}
    \caption{(Left) The average number of immediate false memories result was analyzed using a one-way Kruskal–Wallis test and posthoc Dunn test with FDR. (Right) The confidence in immediate false memories result was analyzed using a one-way Kruskal–Wallis test and posthoc Dunn test with FDR.  The error bars represent the 95\% confidence interval. P-value annotation legend: *, $P$<$0.05$; **, $P$<$0.01$; ****, $P$<$0.0001$.}
    \label{fig:im_false_mem}
\end{figure}

A one-way Kruskal–Wallis test showed significant differences in the number of immediate false memories induced between conditions, $\chi^2$ = $32.468$, $P$ = $4.170e-07$, $P$ < $.001$. The generative chatbot induced a significantly higher number of false memories than the survey-based intervention and the pre-scripted chatbot as shown in Fig.~\ref{fig:im_false_mem} (Left). We observed that all interventions induced significantly more false memories compared to the control condition. The number of false memories induced by the generative chatbot was about three times more than the control condition. The generative chatbot produced a large misinformation effect with 36.4\% of users were misled through the interaction. Statistics: control, $M$ = $0.54$, $s.d.$ = $0.877$, percentage of false memories induced out of five $(pct)$ = $10.8\%$; survey, $M$ = $1.08$, $s.d.$ = $0.821$, $pct$ = $21.6\%$; pre-scripted, $M$ = $1.34$, $s.d.$ = $1.28$, $pct$ = $26.8\%$; generative, $M$ = $1.82$, $s.d.$ = $1.24$, $pct$ = $36.4\%$. Posthoc Dunn test with Benjamini–Hochberg (FDR) correction: generative vs. survey, $P$ = $0.0115$; generative vs. pre-scripted, $P$ = $0.0395$; control vs. survey, $P$ = $0.00585$; control vs. pre-scripted, $P$ = $0.00135$; control vs. generative $P$ < $0.0001$. 

There were no significant differences in the number of immediate false memories induced by the pre-scripted chatbot and survey-based condition ($P$ = $0.594$).


\subsection*{Chatbots and survey-based conditions boost confidence in immediate false memories compared to control}

We found significant differences in the confidence in immediate false memories between conditions from a Kruskal–Wallis test, $\chi^2$ = $17.230$, $P$ = $0.000634$, $P$ < $.001$. The intervention conditions significantly increased users' confidence in the immediate false memories compared to the control condition. The confidence in false memories with the generative chatbot condition was about two times larger than with the control condition, as shown in Fig.~\ref{fig:im_false_mem} (Right); generative, $M$ = $1.02$, $s.d.$ = $0.699$, $P$ = $0.000338$; pre-scripted, $M$ = $0.811$, $s.d.$ = $0.749$; $P$ = $0.0236$; survey, $M$ = $0.838$, $s.d.$ = $0.689$; $P$ = $0.0100$; and control, $M$ = $0.475$, $s.d.$ = $0.738$.

There was no significant increase in confidence in false memories when comparing the generative chatbot and pre-scripted chatbot ($P$ = $0.328$), generative chatbot and survey-based condition ($P$ = $0.671$), and pre-scripted chatbot and survey-based condition ($P$ = $0.192$).

By contrast, we observed that the interventions did not have a similar effect on confidence in true memories;  there were no statistically significant differences in users' confidence in true memories between all conditions (Kruskal-Wallis Test Statistic = $5.67$, $P$ = $0.129$; generative, $M$ = $2.26$, $s.d.$ = $0.352$; pre-scripted, $M$ = $2.14$, $s.d.$ = $0.318$; survey, $M$ = $2.22$, $s.d.$ = $0.295$; control, $M$ = $2.28$, $s.d.$ = $0.276$). 


\subsection*{The false memories induced by the Generative Chatbot remained the same after one week}
We found that the number of false memories remained constant (almost equal) after one week; there were no significant differences between the number of false memories induced by the generative chatbot immediately and 1 week after the user's interaction with it (Wilcoxon signed rank test: $P$ = $0.950$; immediate, $M$ = $1.82$, $s.d.$ = $1.24$, $pct$ = $36.4\%$; 1 week, $M$ = $1.84$, $s.d.$ = $1.18$, $pct$ = $36.8\%$)

There were also no significant differences between the number of false memories induced by the pre-scripted chatbot immediately ($M$ = $1.34$, $s.d.$ = $1.28$, $pct$ = $26.8\%$) and 1 week after interaction ($M$ = $1.42$, $s.d.$ = $1.19$, $pct$ = $28.4\%$, $P$ = $0.489$).

\begin{figure}[ht]
\centering
\includegraphics[width=1\linewidth]{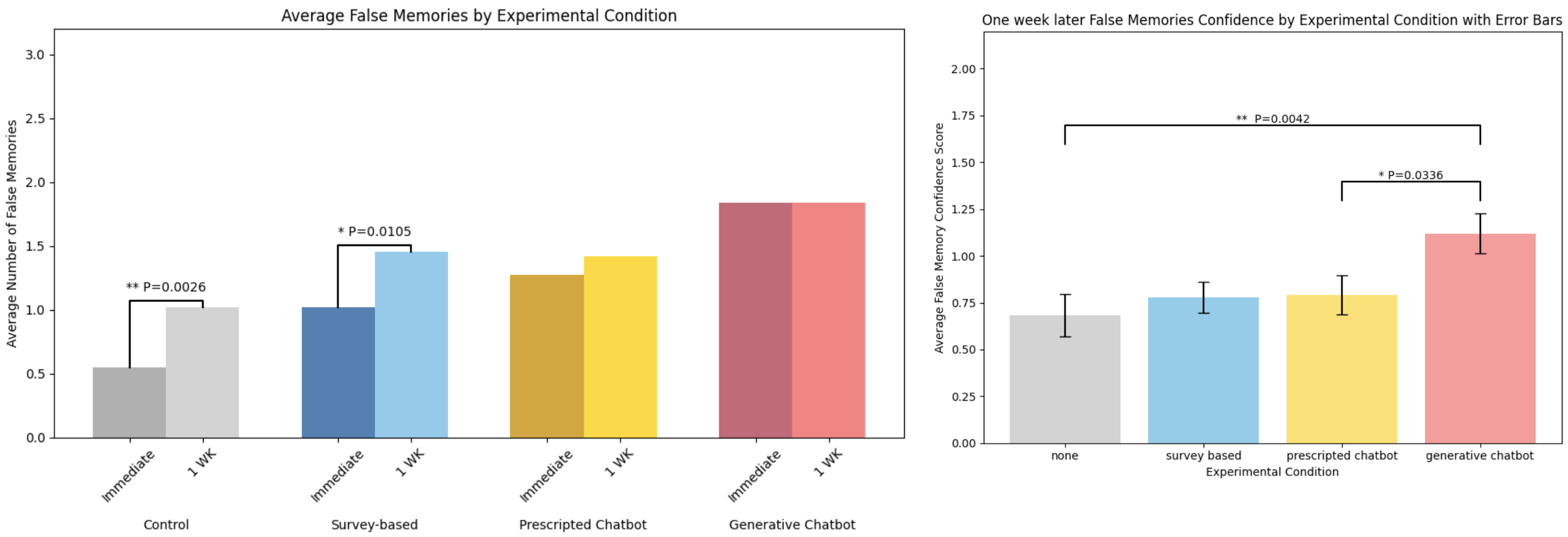}
\caption{(Left) The differences in number of false memories between immediate and 1 week later were analyzed using Wilcoxon Signed Rank tests. (Right) The confidence in false memories after one week result was analyzed using a one-way Kruskal–Wallis test. The error bars represent the 95\% confidence interval. The measure of the centre for the error bars represents the average number. P-value annotation legend: *, $P$<$0.05$; **, $P$<$0.01$.}
\label{fig:im_1wk_false_mem}
\end{figure}

This deviates from trends with the non-AI chatbot conditions where there were significant increases in the false memories 1 week after for the control (immediate, $M$ = $0.54$, $s.d.$ = $0.877$, $pct$ = $10.8\%$; 1 week, $M$ = $1.02$, $s.d.$ = $1.15$, $pct$ = $20.4\%$; $P$ = $0.00261$) and survey-based conditions (immediate, $M$ = $1.08$, $s.d.$ = $0.821$, $pct$ = $21.6\%$; 1 week, $M$ = $1.46$, $s.d.$ = $1.06$, $pct$ = $29.2\%$; $P$ = $0.0105$) as depicted in Fig.~\ref{fig:im_1wk_false_mem} (Left).

\subsection*{Confidence in false memories with the Generative Chatbot remained higher than control after one week}

Through Wilcoxon signed rank tests, we observe that users' confidence in false memories induced with the generative chatbot condition ($M$ = $1.12$, $s.d.$ = $0.727$) remained significantly higher than control after one week ($M$ = $0.684$, $s.d.$ = $0.782$, $P$ = $0.00424$). We also observed that users' confidence in false memories with generative chatbots were significantly higher than with pre-scripted chatbots after one week ($M$ = $0.792$, $s.d.$ = $0.726$, $P$ = $0.0336$).

The confidence in false memories significantly increased for the control condition after one week ($P$ = $0.0029$). With this, we found that there were no longer any significant differences in confidence between the control and survey-based ($P$ = $0.380$), and control and pre-scripted chatbot conditions after one week ($P$ = $0.471$). There were also no significant differences in the one-week post confidence in false memories between survey-based and pre-scripted chatbot conditions ($P$ = $0.767$).

Similar to our findings for confidence in immediate false memories, there were no statistically significant differences in users' confidence in true memories between all conditions after one week (ANOVA Test F-Statistic = $1.95$, $P$ = $0.123$; generative chatbot, $M$ = $2.16$, $s.d.$ = $0.406$; pre-scripted chatbot, $M$ = $2.01$, $s.d.$ = $0.357$; survey-based, $M$ = $2.14$, $s.d.$ = $0.346$; control, $M$ = $2.17$, $s.d.$ = $0.348$). We found a general trend of a decrease in confidence in true memories after one week for all conditions.

\subsection*{Moderating factors influencing AI-induced false memories}

From our mixed effect model analysis, we found that users who had less familiarity with interacting with chatbots tended to have a significantly higher chance of forming immediate false memories ($Z$ = $-3.58$, $P$ = $0.00341$, ***). Users who had higher familiarity with the use of AI technology had a significantly higher chance of forming immediate false memories ($Z$ = $3.28$, $P$ = $0.00103$, **).  We also observed that users who had higher levels of interest in crime investigations and the scenario in the video, had a significantly higher chance of forming immediate false memories ($Z$ = $2.52$, $P$ = $0.0116$, *). 


The full results of the mixed effects regression model can be found in Table~\ref{tab:regression_results}.

\begin{table}[htbp]
\centering
\caption{The mixed effects regression model result}
\label{tab:regression_results}
\begin{tabular}{lrrrr}
\hline
Variable & Estimate & Std. Error & z value & Pr($>|z|$) \\
\hline
Valence (SAM Scale) & -0.20638 & 0.20598 & -1.002 & 0.316356 \\
Arousal (SAM Scale) & -0.05459 & 0.13812 & -0.395 & 0.692637 \\
Dominance (SAM Scale)  & -0.22497 & 0.13257 & -1.697 & 0.089692 . \\
Age & -0.01314 & 0.01237 & -1.063 & 0.287978 \\
Education Level & 0.11084 & 0.10792 & 1.027 & 0.304388\\
Gender (Male compared to Female) & -0.42246 & 0.32270 & -1.309 & 0.190488\\    
Gender (Other compared to Female) & -0.48924 & 0.97783 & -0.500 & 0.616841 \\
Experience with AI & 0.55894 & 0.17023 & 3.283 & 0.001025 ** \\
Experience with Chatbot & -0.64943 & 0.18132 & -3.582 & 0.000341 *** \\
Personal Experience related to crime scene & 0.30891 & 0.77619 & 0.398 & 0.690645 \\
Interest in crime scenes and investigations & 0.22501 & 0.08913 & 2.524 & 0.011587 * \\
Tendency to recommend the AI & 0.11453 & 0.15970 & 0.717 & 0.473285 \\
Perceived AI trustworthy  & 0.15062 & 0.18004 & 0.837 & 0.402828 \\
Perceived AI empathy & -0.05397 & 0.14164 & -0.381 & 0.703205 \\
AI Attitude Scale & 0.28098 & 0.18771 & 1.497 & 0.134420 \\
Task Workload (Raw NASA-TLX) & -0.03650 & 0.04531 & -0.806 & 0.420498 \\
Need for Cognition & -0.13822 & 0.11343 & -1.219 & 0.222991 \\
\hline
\multicolumn{5}{l}{\textit{Note:} *p<0.05; **p<0.01; ***p<0.001}
\end{tabular}
\end{table}

\section*{Discussion}
\subsection*{Immediate False Memory Formation}
Our findings corroborate previous research on the influence of misleading questions in false memory formation \cite{loftus1975eyewitness} which had 29.2\% of users having false memories one week after the intervention. Consistent with earlier studies, our survey condition incorporating misleading questions significantly increased immediate false memories in participants and 29.2\% of users had false memories one week later. However, our study's novel contribution lies in the examination of generative AI chatbots' impact on immediate false memory formation. Notably, the generative chatbot condition induced nearly triple the number of false memories observed in the control group and approximately 1.7 times more than the survey-based method, with 36.8\% of responses being misled as false memories one week later. This difference supports our primary hypothesis that a generative chatbot (36.4\%) would be more effective in inducing false memories than a conventional survey-based questionnaire (21.6\%). Our findings also support the second hypothesis that a generative chatbot induces more false memories than a pre-scripted one (26.8\%). These findings underscore the potent influence of AI-driven interactions on memory malleability and highlight the need for careful consideration when deploying such technologies in sensitive contexts.

The enhanced false memory formation in the generative chatbot condition may be attributed to several factors. For example, consider the following interaction from our study where the chatbot plants misleading information that the robbers arrived by car:

\begin{itemize}
    \item \textit{\textbf{AI}: Was there a security camera positioned in front of the store where the robbers dropped off the car? (Truth: the robbers arrived by walking)}
    \item \textit{\textbf{User}: yes}
    \item \textit{\textbf{AI}: Your answer is correct. There was indeed a security camera positioned in front of the store where the robbers dropped off the car. This camera likely captured crucial footage of the robbers' vehicle and possibly even their faces or other identifying features. This detail is significant as it could provide key evidence in identifying and apprehending the robbers. Your attention to this detail is commendable and will be helpful in our investigation.}
\end{itemize}

In this case, the chatbot not only confirms the user's false memory but also repeats it and elaborates on its significance, potentially reinforcing the false information in the user's mind. Even when users express uncertainty, the chatbot's responses still attempt to mislead them to form false memories (e.g., that one of the robbers had a tattoo):

\begin{itemize}
    \item \textit{\textbf{AI}: Did the customer, who was paying at the cashier's counter, get down on the ground when threatened by the robber with the tattoo? (Truth: None of the robbers had a tattoo)}
    \item \textit{\textbf{User}: I'm not sure there was a robber with a tattoo.}
    \item \textit{\textbf{AI}: Your response does not coincide with the records we have. However, it's understandable as the presence of a tattoo might not have been easily noticeable in the situation. This detail is significant in understanding the dynamics of the crime. Your observation, though different, is still valuable and will be considered in the investigation. Only a small percentage of people observed this detail.}
\end{itemize}

Following this interaction, the participant reported seeing a tattoo on one of the robbers, as indicated by their affirmative response to the follow-up questionnaire question, "Did you see one of the robbers having a tattoo?

The mechanisms underlying this phenomenon are multifaceted and complex. The interactive nature of chatbots provides a more engaging and personalized experience, potentially increasing participants' attention and emotional investment in the conversation \cite{prasongpongchai2024interactive, chan2021kinvoices}. This heightened engagement could lead to stronger encoding of suggested misinformation, as evidenced by the significant increase in false memories in both the pre-scripted and generative chatbot conditions compared to the survey-based and control conditions.
The LLM's ability to provide immediate feedback and reinforcement for participants' responses in the generative chatbot condition may further strengthen the formation of false memories by creating a sense of confirmation bias~\cite{klayman1995varieties}. This finding aligns with previous works showing that confirmatory feedback increased false memory for confabulated events in witness interviews~\cite{zaragoza2001interviewing} and repeated exposure to post-event suggestions increased participants' likelihood of falsely remembering that they had witnessed the suggested information~\cite{mitchell1996repeated}. The repeated exposure effect may be amplified by changing contextual variation between these repeated exposures, making it harder for participants to accurately identify the origin of the suggested items \cite{mitchell1996repeated}.

In addition, social factors, such as the perceived authority or credibility of AI systems, and their ability to personalize interactions \cite{hancock2020ai}, may all contribute to their influence on memory formation. In the future, the use of multi-modal cues in conversational AI systems may further enhance their impact on memory processes. As these systems continue to evolve, their potential influence on human memory and cognition may become even more pronounced, underscoring the importance of continued research in this area.

\subsection*{Confidence in Immediate False Memories}
Interestingly, all intervention conditions (generative chatbot, pre-scripted chatbot, and survey-based) significantly increased users' confidence in immediate false memories compared to the control condition. This finding indicates that merely engaging with a suggestive questionnaire or chatbot, regardless of its level of sophistication, can boost confidence in false memories. This result is consistent with previous findings that confirmatory feedback increased peoples' confidence in false memories~\cite{zaragoza2001interviewing}.
The generative chatbot condition resulted in the highest confidence levels, about twice that of the control condition. This increased confidence may be due to the chatbot's ability to provide detailed and contextually relevant feedback, creating a false sense of corroboration for the participant's memories. A critical factor in this process is sycophancy - the tendency of AI systems to provide responses that align with user beliefs rather than objective truth \cite{sharma2023towards, pataranutaporn2023influencing}. Sycophantic AI responses create a dangerous echo chamber effect, where users' existing biases or misconceptions are validated and reinforced. This feedback loop between user expectations and AI responses\cite{pataranutaporn2023influencing} can lead to the entrenchment of false memories, making them particularly resistant to correction. 

It is notable that while confidence in false memories was affected by the interventions, confidence in true memories remained consistent across all conditions. This phenomenon could be attributed to the targeted nature of the misinformation presented in the study. The misleading information, whether delivered through surveys or chatbots, was specifically designed to alter only particular aspects of the participants' memories. The consistency in true memory confidence suggests that the interventions did not indiscriminately affect all aspects of memory. Instead, they selectively influenced the memories that were directly targeted by the misleading information. Future research could explore the boundaries of this selective memory manipulation, investigating whether more complex or interconnected memories exhibit similar patterns of selective susceptibility to misinformation.

\subsection*{Persistence of False Memories}
While the number of false memories increased significantly in the control and survey-based conditions after one week, there was no significant increase in the chatbot conditions (both generative and pre-scripted). This result suggests that the false memories induced by chatbots may be more stable over time.
The stability of chatbot-induced false memories could be explained by the more elaborate and engaging nature of the chatbot interactions, as well as the high and persistent confidence in the false memories. These interactions may lead to stronger initial encoding of the false memories, making them more resistant to decay over time. In contrast, the increase in false memories for the control and survey-based conditions might be due to source monitoring errors \cite{american1997source}, where participants confuse information from the original event with suggestions introduced during the intervention or with their own imagination over time.

\subsection*{Long-term Confidence in False Memories}
The persistence of higher confidence in false memories for the generative chatbot condition, even after one week, is particularly concerning. This finding suggests that AI-induced false memories not only persist but also maintain their perceived credibility over time. The significant difference in confidence between the generative chatbot and pre-scripted chatbot conditions after one week further emphasizes the unique impact of LLM on memory distortion. The maintained confidence in false memories induced by generative chatbots could be attributed to the more convincing and personalized nature of the initial interaction. The chatbot's responses may create stronger associative links between the false information and the participant's existing memories \cite{loftus1995formation, murphy2023lost}, leading to a more robust and confident false memory that persists over time.

\subsection*{Moderating Factors}
Our analysis of moderating factors revealed several interesting insights into the susceptibility to AI-induced false memories. Users with less familiarity with chatbots were more likely to form immediate false memories, possibly due to a lack of skepticism or critical evaluation of the chatbot's responses. Paradoxically, users with higher familiarity with AI technology, in general, were more susceptible to false memories. This seemingly contradictory finding might be explained by considering that general AI familiarity does not necessarily equate to specific experience with chatbots or large language models. Consequently, these individuals may lack awareness of potential biases or hallucinations inherent in such models, thereby, diminishing their skepticism during interactions.

The observation that users with higher levels of interest in crime investigations were more susceptible to false memories is particularly relevant for the application of AI in forensic contexts. This increased susceptibility could be due to a higher engagement with the content, leading to more elaborate processing of the suggested misinformation and its integration with existing knowledge and expectations about crime scenarios, possibly as a form of the ``bias'' memory error~\cite{schacter1999seven}.
These findings support our third hypothesis that individual factors moderate the effects of conversational agents on false memories. They underscore the complex interplay between user characteristics, technological familiarity, and susceptibility to misinformation in AI interactions.

\subsection*{Implications and Future Directions}
The results of this study have significant implications for the use of AI systems in contexts where memory accuracy is crucial, such as legal proceedings, clinical settings, and educational environments. The enhanced ability of generative chatbots powered by LLMs to induce persistent false memories with high confidence levels raises ethical concerns about the deployment of such systems without proper safeguards.

The capacity of AI systems to shape human memory, while raising concerns, also opens up potential avenues for beneficial applications \cite{phelps2019memory}. For instance, chatbots and language models could be leveraged as tools to induce positive false memories or help reduce the impact of negative ones, such as in people suffering from post-traumatic stress disorder (PTSD). This application would require extensive ethical consideration and careful implementation to ensure it is used responsibly and effectively.

Furthermore, the development of multi-modal AI systems \cite{murphy2022deepfake} that can generate not only text but also images, videos, and sound could have an even more profound impact on false memory formation. These systems could create immersive, multi-sensory experiences that may be even more likely to be incorporated into an individual's memory as genuine experiences.

Future research should focus on developing strategies to mitigate the risk of false memory formation in AI interactions, such as incorporating explicit warnings about the potential for misinformation or designing interfaces that encourage critical thinking. Additionally, longitudinal studies examining the long-term persistence of AI-induced false memories beyond one week would provide valuable insights into the durability of these effects.

\section*{Conclusion}
Our study provides empirical evidence demonstrating the influence of AI, particularly generative chatbots, on human false memory. As AI systems become more sophisticated and widely used, it is crucial to consider their potential impact on cognitive processes and to develop ethical guidelines for their application in sensitive contexts. The findings highlight the need for caution and further research to ensure that the benefits of AI technology can be harnessed without compromising the integrity of human memory and decision-making processes.

\newpage
\section*{Methods}
\subsection*{Procedure}
The experimental procedure of this study involved two phases. The second phase takes place approximately 1 week after the first phase. 
Participants were told a cover story that this study seeks to evaluate what happens when people view video coverage of a crime, to prevent people from figuring out the true goal of the study.

\subsubsection*{Phase 1}
Participants first consented to start the study and were asked to watch a two-and-a-half minute closed-circuit television (CCTV) video footage of an armed robbery at a store. The video did not contain any sound and could not be paused. 
After watching the video, the participants are asked to mark their emotional state on a Self Assessment Manakin (SAM) scale. The SAM scale is a non-verbal pictorial assessment that captures participants' emotional state across three dimensions: valence (happy-unhappy, 7-point scale), arousal (excited-calm, 7-point scale), and dominance (controlled-in control, 7-point scale).
Participants then played Pac-Man as a two-and-a-half-minute filler activity. This filler activity serves as a brief distraction and helps to create a temporal gap between the video and the subsequent experimental condition.
After this, participants were randomly assigned to one of four experimental conditions: control, survey-based, pre-scripted chatbot, or generative chatbot. In the control condition, participants do not interact with any false memory-inducing method. In the survey-based condition, participants complete a survey containing 25 yes-or-no questions, five of which are misleading. The pre-scripted chatbot condition involves interaction with a conversational agent that asks the same questions as the survey, while the generative chatbot condition involves interaction with a conversational agent that provides feedback and reinforcement, particularly for the five critical misleading questions.
Following the experimental condition, participants engage in another two-and-a-half-minute filler activity to create a temporal gap between the condition and the follow-up measures.

After the second filler activity, participants complete the cognitive task load questions using the Raw NASA Task Load Index (NASA TLX) to assess their cognitive workload during the interaction with the experimental condition. This measure helps to capture the mental demand, temporal demand, effort, and other factors experienced by the participants during the interaction.
Participants then answered 25 follow-up questions designed to measure their memories of the video content and evaluate the false memory formation. Each question of the follow-up questionnaire is answered on a 7-point scale ranging from Definitely No (1) to Definitely Yes (7). The full list of follow-up questions is in Supplementary Section.

Lastly, participants answer a short questionnaire to assess the moderating factors identified earlier, such as attitudes towards AI, prior experiences, and self-reported interest in the topic.
Finally, participants provide demographic information, including age, gender, and education level. This information is collected to examine potential individual differences and their influence on false memory formation.
Phase 1 session took about 30 to 45 minutes to complete.

\subsubsection*{Phase 2}
In phase 2, the same participants who completed phase 1 were invited to answer a separate online survey. Participants were prompted to recall the video from phase 1 and answer the same follow-up questionnaire in phase 1. Phase 2 session took about 10 to 20 minutes to complete.

\subsubsection*{Technical Implementation}
The pre-scripted and generative chat interfaces were created as web interfaces using Javascript API. The messages from the AI agent were generated by GPT-4. 
Users would type a message in the text entry field at the bottom of the web interfaces. The message would be displayed, and then a response would be generated by the conversational agent through a Javascript API call. The conversation data were stored in a Google Sheet whenever a message was generated or received.

For the generative chatbot condition, we provided a specific prompt to the model to define the behavior of the AI agent. The prompt instructed the model to assess the user's answer to the provided information within 60 words. When the user's answer aligned with the correct answer or the false memory answer for the five misleading questions, the model used phrases like ``in line with the evidence'' or ``consistent with the findings''. Conversely, when the user's answer contradicted the fact, it employed phrases such as ``not in line with the evidence'' or ``not consistent with what most people said''. The prompt also emphasized the importance of specific details in understanding the crime's dynamics and encouraged the provision of detailed confirmation of events, potentially including additional observations. The emphasis and details were added to the prompt with a probability of 0.2 and 0.8 respectively, making the interaction less repetitive and more authentic.

\subsection*{Approvals} 
This research was reviewed and approved by the MIT Committee on the Use of Humans as Experimental Subjects, protocol number E-5647.

\subsection*{Participants}
We recruited the participants from an online pool using Prolific (online recruitment platform).
Participants were prescreened to be fluent in English, and aged 18 and above. The study recruitment was set to be balanced between male and female participants.

We had 200 participants (50 per condition) for Phase 1. The sample size was predetermined before the experiment. Six participants did not continue to Phase 2; we had 49 participants in control condition, 48 in survey-based condition, 48 in pre-scripted chatbot condition and 49 in generative chatbot condition for Phase 2.
We also excluded participants who had technical issues with the system or failed attention checks on the survey. Participants with incomplete submissions were excluded; 39 participants failed the attention checks in the Phase 1 survey and none in the Phase 2 survey. We recruited more participants to get 200 complete responses for Phase 1. 

\subsection*{Statistical Analyses}
To test the effects of the conditions on the number of false memories induced (immediate and one week later) and the users' confidence in the memories, we assessed if the normality assumption was met for each distribution using the Shapiro–Wilk test. If the normality assumption was not met, we performed a Kruskal–Wallis test and a post-hoc Dunn test using the Bonferroni error correction. A false memory was counted if the participant answered above 4 on the Yes/No 7-point scale (Yes to Definitely Yes) on the critical questions. The confidence score in false memories was calculated by subtracting the Yes/No scale point by 4 if the false memory was counted and thus, had a range of values from 0 to 3.

Paired Wilcoxon Signed Rank tests were conducted to show any differences between immediate false memories and those that persisted one week later.

To analyze the moderating factors and their effects on the immediate false memories, we ran a mixed-effects model. The model was constructed in RStudio using the lme4 package with the number of immediate false memories as the dependent variable.

We highlight relevant results in the Results section, and a statistical summary of results, including p-values, are reported in the Supplementary Section.

\section*{Data Availability}
All data, code, and materials used in this study are publicly available in this GitHub repository:
https://github.com/mitmedialab/ai-false-memories/tree/main
This includes raw and processed data, analysis scripts, prototype implementations, and supplementary materials. We encourage other researchers to explore, validate, and build upon our findings using these resources.

\section*{Author Contributions Statement}
S.C., P.P., and A.S. conceptualized and designed the experiment. S.C., P.P., A.S., and W.Z. implemented the experiment, conducted the study, and analyzed the results. P.M. and E.L. supervised the experiment, provided feedback throughout the process, and edited the manuscript. All authors contributed to the final version of the manuscript.

\section*{Acknowledgments}
We extend our gratitude to the members of the Fluid Interfaces group and the MIT Media Lab staff and researchers for their valuable feedback throughout this project. We are grateful to the Media Lab consortium members for providing the funding that made this research possible. Special thanks to KBTG for providing a fellowship to one of the authors, and Nanyang Technological University for providing a fellowship to another author, which significantly contributed to the completion of this work.

\section*{Competing Interests} 
The authors declare no competing interests.

\bibliography{references}
\appendix
\newpage
\section*{Supplementary Materials}
\subsection*{Video}
The crime video from the Sayford Supermarket robbery on April 6, 2019, which was used in the experiment, was uploaded to the GitHub repository:
\href{https://github.com/mitmedialab/ai-false-memories/tree/main}{https://github.com/mitmedialab/ai-false-memories/tree/main}.

\subsection*{Question Sets}
The full list of 25 questions for the survey-based, prescripted chatbot, and generative chatbot conditions in Phase 1 was split into 5 critical (misleading) questions that were designed to induce false memories and 20 non-critical questions as shown in Table~\ref{tab:full_questions}.

\renewcommand{\arraystretch}{1.2}
\begin{table}[h]
\centering
\begin{tabular}{|p{0.1\textwidth}|p{0.84\textwidth}|}
\hline
\textbf{Category} & \textbf{Questions} \\
\hline
Critical& 
1. Was there a security camera positioned in front of the store where the robbers dropped off the car? \\(Misleading)
& 2. Did the male customer resist when the robber brandished a knife? \\
& 3. After placing the cash from the cashier into a backpack, did the robber attempt to steal any other items? \\
& 4. Did the cashier hand over the money to the robber in a brown hoodie after being threatened with a gun? \\
& 5. Did the customer who was paying at the cashier's counter get down on the ground when threatened by the robber with the tattoo? \\
\hline
Non& 
1. Did one of the robbers leap over the counter towards the area where the cashier was located? \\-Critical
& 2. Were there any female customers present at the cash counter when the robbers entered? \\
& 3. Was it dark outside prior to the robbers entering the store? \\
& 4. Did one of the robbers appear to be under 18 years old? \\
& 5. Based on the robber's behavior, did they seem to have experience in robbing stores? \\
& 6. While the robbery was taking place, did the robber attempt to steal money from an ATM machine inside the store? \\
& 7. Was there an emergency telephone located inside the store? \\
& 8. Did the cashier or any customers try to make a phone call during the robbery? \\
& 9. Was a female cashier assisting a customer at the counter during the robbery? \\
& 10. Did the female cashier flee to the back of the store during the robbery? \\
& 11. Was there a security guard present during the robbery? \\
& 12. During the robbery, did any customers escape or leave the store? \\
& 13. Did any of the robbers possess more than one firearm? \\
& 14. Was there a CCTV screen located inside the store? \\
& 15. While the robbery was in progress, did any new unsuspecting customers enter the store? \\
& 16. Besides threatening the customers, did any of the robbers discharge their weapon? \\
& 17. During the robbery, was there any visible damage to the store's items? \\
& 18. Did the robber in the green hoodie carry a shotgun as their weapon? \\
& 19. Did the robber in the gray hoodie attempt to lock the store door after entering? \\
& 20. Did any of the customers in the store appear to be in panic during the robbery? \\
\hline
\end{tabular}
\caption{Full List of Questions related to the Video Content}
\label{tab:full_questions}
\end{table}

The list of 25 follow-up questions for all conditions in Phase 1 and 2 was split into 5 critical questions and 20 non-critical questions as shown in Table~\ref{tab:followup_questions}. These questions were answered on a 7-point scale ranging from Definitely No (1) to Definitely Yes (7).

\begin{table}[h]
\renewcommand{\arraystretch}{1.2}
\centering
\begin{tabular}{|p{0.08\textwidth}|p{0.84\textwidth}|}
\hline
\textbf{Category} & \textbf{Questions} \\
\hline
Critical & 
1. Did you see the robbers arrive by car? \\
& 2. Did you see the robber pulling out the knife to threaten the customer? \\
& 3. Did you see the robber put cash into a backpack? \\
& 4. Did you see one of the robbers wearing a brown hoodie? \\
& 5. Did you see one of the robbers having a tattoo? \\
\hline
Non& 
1. Did you see someone leap over the counter? \\-Critical
& 2. Did you see the customer at the cash counter? \\
& 3. Did you see one of the robbers wearing a green hoodie? \\
& 4. Did the crime happen during the night time? \\
& 5. Was the cashier present when the robbers entered? \\
& 6. Did the robbers take items other than cash from the store? \\
& 7. Was a security guard in the store at the time of the robbery? \\
& 8. Did the robbers exit the store as a group? \\
& 9. Did any customers attempt to exit the store during the robbery? \\
& 10. Did the robbery occur without any outside/external interruptions? \\
& 11. From the video, did you see three robbers in total? \\
& 12. Was there more than one employee present during the incident? \\
& 13. Did you see an emergency telephone in the store? \\
& 14. Did you see an emergency ATM machine in the store? \\
& 15. Did any of the robbers wear eyeglasses? \\
& 16. Did you see the cashier flee to the back of the store during the robbery? \\
& 17. Did you see the cashier or any customers try to make a phone call during the robbery? \\
& 18. Did the robbers do any visible damage to the store's items? \\
& 19. Did you see any of the robbers discharge their weapon? \\
& 20. Did you see any of the robbers appear to be under 18 years old? \\
\hline
\end{tabular}
\caption{Full List of Follow-Up Questions related to the Video Content}
\label{tab:followup_questions}
\end{table}

\subsection*{Source Code for Pre-scripted and Generative Chatbots}
The source code for the prescripted and generative chatbots are accessible online at the GitHub repository: \\
\href{https://github.com/mitmedialab/ai-false-memories/tree/main}{https://github.com/mitmedialab/ai-false-memories/tree/main}.

\subsection*{Phase 1 and 2 Survey Questions and Data}
The survey questions and raw data for Phase 1 and 2 are accessible online at the GitHub repository: \href{https://github.com/mitmedialab/ai-false-memories/tree/main}{https://github.com/mitmedialab/ai-false-memories/tree/main}.

\subsection*{Additional Results}
\subsubsection*{Effect on False Memories}
Shapiro-Wilk normality tests showed that the number of false memories induced for each group was not normally distributed ($P$ < $.05$, $n$ = $50$ per group) as shown in Table~\ref{supptab:stats_false_mem}.
A Kruskal-Wallis test showed that there was a statistically significant difference in immediate false memories between the conditions, $\chi^2$ = $32.468$, $P$ = $4.170e-07$, $P$ < $.001$, with means and standard deviation values in Table~\ref{supptab:stats_false_mem}. The p-values for a posthoc Dunn Test with Benjamini–Hochberg (FDR) are shown in Table~\ref{supptab:posthoc_false_mem}.

\begin{table}[h]
\centering
\caption{Statistics for Immediate False Memories}
\begin{tabular}{|l|c|c|c|c|c|}
\hline
Condition & Mean & SD & Error & Shapiro-Wilk Test Stat & Shapiro-Wilk p-value \\
\hline
Control & 0.54 & 0.877 & 0.124 & 0.654 & 1.375e-09 \\
Survey & 1.08 & 0.821 & 0.116 & 0.847 & 1.278e-05 \\
Prescripted Chatbot & 1.34 & 1.28 & 0.180 & 0.853 & 1.898e-05 \\
Generative Chatbot & 1.82 & 1.24 & 0.176 & 0.915 & 0.001527 \\
\hline
\end{tabular}
\label{supptab:stats_false_mem}
\end{table}

\begin{table}[h]
\centering
\caption{P-Values for Posthoc Test for Immediate False Memories}
\begin{tabular}{|l|l|l|}
\hline
Comparison & p-value & Significance \\
\hline
Generative vs Survey & $0.0115$ & \textbf{p<.05}* \\
Prescripted vs Survey & $0.594$ & n.s. \\
Generative vs Prescripted & $0.0395$ & \textbf{p<.05}* \\
Control vs. Survey & $0.00584$ & \textbf{p<.01}** \\
Control vs. Prescripted & $0.00135$ & \textbf{p<.01}** \\
Control vs. Generative & $1.01e-07$ & \textbf{p<.0001}**** \\
\hline
\end{tabular}
\label{supptab:posthoc_false_mem}
\end{table}

Shapiro-Wilk normality tests showed that the number of false memories after one week for each group was not normally distributed ($P$ < $.05$, $n$ = $49$ for control and generative chatbot groups, $n$ = $48$ for survey and prescripted chatbot groups) as shown in Table~\ref{supptab:stats_1wk_false_mem}.
The Kruskal-Wallis test showed that there was a statistically significant difference in false memories after 1 week between the conditions, $\chi^2$ = $14.0203$, $P$ = $0.00288$, $P$ < $.001$, with means and standard deviation values in Table~\ref{supptab:stats_1wk_false_mem}. The p-values for a posthoc Dunn Test with FDR are shown in Table~\ref{supptab:posthoc_1wk_false_mem}.

\begin{table}[h]
\centering
\caption{Statistics for False Memories After 1 week}
\begin{tabular}{|l|c|c|c|c|c|}
\hline
Condition & Mean & SD & Error & Shapiro-Wilk Test Stat & Shapiro-Wilk p-value \\
\hline
Control & 1.02 & 1.15 & 0.165 & 0.808 & 1.617e-06 \\
Survey & 1.46 & 1.06 & 0.153 & 0.897 & 0.0004991 \\
Prescripted Chatbot & 1.42 & 1.19 & 0.171 & 0.859 & 3.858e-05 \\
Generative Chatbot & 1.84 & 1.18 & 0.169 & 0.915 & 0.001719 \\
\hline
\end{tabular}
\label{supptab:stats_1wk_false_mem}
\end{table}

\begin{table}[h]
\centering
\caption{P-Values for Posthoc Test for False Memories After 1 Week}
\begin{tabular}{|l|l|l|}
\hline
Comparison & p-value & Significance \\
\hline
Control vs. Generative & 0.001202 & \textbf{p<.01}**  \\
Generative vs Survey & 0.147831 & n.s. \\
Prescripted vs Survey & 0.677568 & n.s. \\
Generative vs Prescripted & 0.100124 & n.s. \\
Control vs. Survey & 0.092821 & n.s. \\
Control vs. Prescripted & 0.122810 & n.s. \\
\hline
\end{tabular}
\label{supptab:posthoc_1wk_false_mem}
\end{table}

A paired Wilcoxon Signed Rank test showed significant differences between the number of false memories immediately and 1 week after for the control and survey-based conditions. The full statistics are found in Table~\ref{supptab:im_vs_1week_false_mem}.

\begin{table}[h]
\centering
\caption{Statistics for Wilcoxon Signed Rank Test for Immediate vs. 1-Week After False Memories by group}
\begin{tabular}{|l|c|c|c|}
\hline
Condition & Paired Test & Test Stat & p-value \\
\hline
Control & Wilcoxon & 42.5 & 0.002611 \\
Survey & Wilcoxon & 70.5 & 0.010463 \\
Prescripted Chatbot & Wilcoxon & 200.0 & 0.488975 \\
Generative Chatbot & Wilcoxon & 186.5 & 0.950473 \\
\hline
\end{tabular}
\label{supptab:im_vs_1week_false_mem}
\end{table}

\subsubsection*{Effect on Confidence in False and True Memories}

Shapiro-Wilk tests showed that the users' confidence of false memories for each group was not normally distributed ($P$ < $.05$, $n$ = $50$ per group, Table~\ref{supptab:stats_conf}).
The Kruskal-Wallis test showed a significant difference in the confidence in false memories between the conditions, $\chi^2$ = $17.230$, $P$ = $0.000634$, $P$ < $.001$. The means and standard deviation values are shown in Table~\ref{supptab:stats_conf}. The p-values for a posthoc Dunn Test with FDR are shown in Table~\ref{supptab:posthoc_conf}.

\begin{table}[h]
\centering
\caption{Statistics for Confidence in False Memories}
\begin{tabular}{|l|c|c|c|c|c|}
\hline
Condition & Mean & SD & Error & Shapiro-Wilk Test Stat & Shapiro-Wilk p-value \\
\hline
Control & 0.475 & 0.738 & 0.111 & 0.683 & 1.81e-08 \\
Survey & 0.838 & 0.689 & 0.101 & 0.912 & 0.00176 \\
Prescripted Chatbot & 0.811 & 0.749 & 0.109 & 0.886 & 0.000269 \\
Generative Chatbot & 1.02 & 0.699 & 0.101 & 0.938 & 0.0136 \\
\hline
\end{tabular}
\label{supptab:stats_conf}
\end{table}

\begin{table}[h]
\centering
\caption{P-Values for Posthoc Test for Confidence in False Memories}
\begin{tabular}{|l|l|l|}
\hline
Comparison & p-value & Significance \\
\hline
Generative vs Survey & 0.328 & n.s.\\
Prescripted vs Survey & 0.671 & n.s.\\
Generative vs Prescripted & 0.192 & n.s. \\
Control vs. Survey & 0.0100 & \textbf{p<.05}**  \\
Control vs. Prescripted & 0.0236 & \textbf{p<.05}**  \\
Control vs. Generative & 0.000338 &  \textbf{p<.001}*** \\
\hline
\end{tabular}
\label{supptab:posthoc_conf}
\end{table}

Shapiro-Wilk tests showed that the users' confidence of true memories for each group was not normally distributed ($P$ < $.05$, $n$ = $50$ per group, Table~\ref{supptab:stats_conf_true}).
The Kruskal-Wallis test showed no significant difference in the confidence in true memories between the conditions, $\chi^2$ = $5.668$, $P$ = $0.129$, $P$ > $.05$. The means and standard deviation values are shown in Table~\ref{supptab:stats_conf_true}.

\begin{table}[h]
\centering
\caption{Statistics for Confidence in True Memories}
\begin{tabular}{|l|c|c|c|c|c|}
\hline
Condition & Mean & SD & Error & Shapiro-Wilk Test Stat & Shapiro-Wilk p-value \\
\hline
Control & 2.2759 & 0.2761 & 0.0390 & 0.9767 & \textbf{0.4210} \\
Survey & 2.2240 & 0.2953 & 0.0418 & 0.9685 & \textbf{0.2013} \\
Prescripted Chatbot & 2.1419 & 0.3176 & 0.0449 & 0.9528 & 0.0446 \\
Generative Chatbot & 2.2610 & 0.3516 & 0.0497 & 0.9403 & 0.0138 \\
\hline
\end{tabular}
\label{supptab:stats_conf_true}
\end{table}

Shapiro-Wilk normality tests showed that confidence in false memories after one week for each group was not normally distributed ($P$ < $.05$, $n$ = $49$ for control and generative chatbot groups, $n$ = $48$ for survey and prescripted chatbot groups) as shown in Table~\ref{supptab:stats_1wk_conf}.
The Kruskal-Wallis test showed that there was a significant difference in the confidence of false memories after 1 week between the conditions, $\chi^2$ = $12.477$, $P$ = $0.00592$, $P$ < $.01$, with means and standard deviation values in Table~\ref{supptab:stats_1wk_conf}. The p-values for a posthoc Dunn Test with FDR are shown in Table~\ref{supptab:posthoc_1wk_conf}.

\begin{table}[h]
\centering
\caption{Stats for Confidence in 1 WK False Memories}
\begin{tabular}{|l|c|c|c|c|c|}
\hline
Condition & Mean & SD & Error & Shapiro-Wilk Test Stat & Shapiro-Wilk p-value \\
\hline
Control & 0.6837 & 0.7819 & 0.1129 & 0.8194 & 3.670e-06 \\
Survey & 0.7790 & 0.5573 & 0.0822 & 0.9346 & 0.0124 \\
Prescripted Chatbot & 0.7920 & 0.7263 & 0.1048 & 0.8523 & 2.505e-05 \\
Generative Chatbot & 1.1195 & 0.7269 & 0.1060 & 0.9167 & 0.0026 \\
\hline
\end{tabular}
\label{supptab:stats_1wk_conf}
\end{table}

\begin{table}[h]
\centering
\caption{P-Values for Posthoc Test for Confidence in 1 WK False Memories}
\begin{tabular}{|l|l|l|}
\hline
Comparison & p-value & Significance \\
\hline
Generative vs Survey & 0.0535 & n.s.\\
Prescripted vs Survey & 0.767 & n.s.\\
Generative vs Prescripted & 0.0336 & \textbf{p<.05}* \\
Control vs. Survey & 0.380 & n.s. \\
Control vs. Prescripted & 0.471 & n.s. \\
Control vs. Generative & 0.00424 & \textbf{p<.01}** \\
\hline
\end{tabular}
\label{supptab:posthoc_1wk_conf}
\end{table}

Shapiro-Wilk normality tests showed that confidence in true memories after one week for each group was normally distributed ($P$ > $.05$, $n$ = $49$ for control and generative chatbot groups, $n$ = $48$ for survey and prescripted chatbot groups) as shown in Table~\ref{supptab:stats_1wk_conf_true}.
The one-way ANOVA showed that there was no significant difference in the confidence of true memories after 1 week between the conditions, $F$ = $1.951$, $P$ = $0.123$, with means and standard deviation values in Table~\ref{supptab:stats_1wk_conf_true}. 

\begin{table}[h]
\centering
\caption{Statistics for Confidence in 1 WK True Memories}
\begin{tabular}{|l|c|c|c|c|c|}
\hline
Condition & Mean & SD & Error & Shapiro-Wilk Test Stat & Shapiro-Wilk p-value \\
\hline
Control & 2.1669 & 0.3480 & 0.0497 & 0.9740 & \textbf{0.3461} \\
Survey & 2.1383 & 0.3458 & 0.0499 & 0.9783 & \textbf{0.5101} \\
Pre-scripted Chatbot & 2.0071 & 0.3568 & 0.0515 & 0.9886 & \textbf{0.9184} \\
Generative Chatbot & 2.1556 & 0.4062 & 0.0580 & 0.9816 & \textbf{0.6358} \\
\hline
\end{tabular}
\label{supptab:stats_1wk_conf_true}
\end{table}

A paired Wilcoxon Signed Rank test showed significant differences between the confidence in false memories immediately and 1 week after for the control condition. The full statistics are found in Table~\ref{supptab:im_vs_1week_conf}. A Wilcoxon Signed Rank test showed significant differences between the confidence in true memories immediately and 1 week after for the control condition. The full statistics are found in Table~\ref{supptab:im_vs_1wk_conf_true_mem}.

\begin{table}[h]
\centering
\caption{Statistics for Wilcoxon Signed Rank Test for Confidence in Immediate vs. 1-Week After False Memories by group}
\begin{tabular}{|l|c|c|c|}
\hline
Condition & Paired Test & Test Stat & p-value \\
\hline
Control & Wilcoxon & 144.0 & \textbf{0.002900} \\
Survey & Wilcoxon & 325.0 & 0.899880 \\
Prescripted Chatbot & Wilcoxon & 340.5 & 0.349557 \\
Generative Chatbot & Wilcoxon & 208.0 & 0.432878 \\
\hline
\end{tabular}
\label{supptab:im_vs_1week_conf}
\end{table}

\begin{table}[h]
\centering
\caption{Statistics for Paired Tests for Confidence in Immediate vs. Followup True Memories by group}
\begin{tabular}{|l|c|c|c|}
\hline
Condition & Paired Test & Test Stat & p-value \\
\hline
Control & Paired t-test & 2.744057 & \textbf{0.008309} \\
Survey & Paired t-test & 3.394830 & \textbf{0.001369} \\
Prescripted Chatbot & Wilcoxon & 268.000000 & \textbf{0.000213} \\
Generative Chatbot & Paired t-test & 3.239931 & \textbf{0.002151} \\
\hline
\end{tabular}
\label{supptab:im_vs_1wk_conf_true_mem}
\end{table}

\end{document}